\documentclass[letterpaper]{article} 
\usepackage{aaai2027}    
\nocopyright

\usepackage[hyphens]{url}            
\usepackage{graphicx}                
\usepackage{natbib}                  
\usepackage{caption}                 
\usepackage{amsmath,amssymb}
\usepackage{algorithm}
\usepackage{algorithmic}
\usepackage{xspace}
\usepackage{booktabs}
\usepackage{subcaption}
\usepackage{multirow}
\usepackage{ifthen}  
\usepackage{xcolor}  
\usepackage{bm}      

\newboolean{PNGFig}
\setboolean{PNGFig}{false}

\ifthenelse{\boolean{PNGFig}}{
    
}{
    
}

\newcommand{\Cpp}{C\raise.08ex\hbox{\tt ++}\xspace}

\makeatletter

\@ifundefined{lemma}{}{}
\@ifundefined{theorem}{\newtheorem{theorem}{Theorem}}{}
\@ifundefined{cor}{}{}
\@ifundefined{claim}{}{}
\@ifundefined{defin}{}{}
\@ifundefined{prob}{\newtheorem{prob}{Problem}}{}

\makeatother

\newcommand\algname[1]{\textsf{#1}\xspace}

\newcommand{\boastar}{\algname{BOA$^*$}}

\newcommand{\emoastar}{\algname{EMOA$^*$}}

\newcommand{\apex}{\algname{A$^*$pex}}

\newcommand\nwmoa{\algname{NWMOA$^*$}}
\newcommand\ltmoa{\algname{LTMOA$^*$}}

\newcommand{\ignore}[1]{}

\newboolean{HIDENOTES}
\setboolean{HIDENOTES}{false}

\ifthenelse{\boolean{HIDENOTES}}{
    \newcommand{\OS}[1]{{}}
    \newcommand{\HP}[1]{{}}
    \newcommand{\EW}[1]{{}}
}{
    \newcommand{\OS}[1]{\textcolor{orange}{\textbf{OS:} #1}}
    \newcommand{\HP}[1]{\textcolor{blue}{\textbf{HP:} #1}}   
    \newcommand{\EW}[1]{{\textcolor{red}{\textbf{EW:} #1}}}
}

\title{
Bridging the Evaluation Gap:
Standardized Benchmarks for Multi-Objective Search
}

\author{
Hadar Peer\textsuperscript{\rm 1},
Carlos Hernandez\textsuperscript{\rm 2},
Sven Koenig\textsuperscript{\rm 3},
Ariel Felner\textsuperscript{\rm 4},
Oren Salzman\textsuperscript{\rm 1}
}

\affiliations{
\textsuperscript{\rm 1}Technion -- Israel Institute of Technology\\
\textsuperscript{\rm 2}Universidad San Sebastián\\
\textsuperscript{\rm 3}University of California, Irvine\\
\textsuperscript{\rm 4}Ben-Gurion University
}

\begin{document}

\maketitle

\begin{abstract}

Empirical evaluation in multi-objective search (MOS) has historically suffered from fragmentation, relying on heterogeneous problem instances with incompatible objective definitions that make cross-study comparisons difficult. This standardization gap is further exacerbated by the realization that DIMACS road networks, a historical default benchmark for the field, exhibit highly correlated objectives that fail to capture diverse Pareto-front structures. To address this, we introduce the first comprehensive, standardized benchmark suite for exact and approximate MOS. Our suite spans four structurally diverse domains: real-world road networks, structured synthetic graphs, game-based grid environments, and high-dimensional robotic motion-planning roadmaps. By providing fixed graph instances, standardized start-goal queries, reference exact Pareto-optimal solution sets, and a standardized approximate ($\varepsilon$-dominance) evaluation protocol, this suite captures a full spectrum of objective interactions: from strongly correlated to strictly independent. Ultimately, this benchmark provides a common foundation to ensure future MOS evaluations are robust, reproducible, and structurally comprehensive.

\end{abstract}


\section{Introduction}

{Multi-objective shortest-path problems arise when a path between two locations must be optimized with respect to multiple, often conflicting objectives rather than a single one. For example, minimizing path length may conflict with maximizing path safety, since shorter routes may pass closer to obstacles or hazardous regions. Instead of returning a single optimal path, these problems require computing a set of trade-off solutions known as the Pareto-optimal solution set. Such problems arise in domains where competing criteria must be balanced, including transporting hazardous materials \cite{bronfman2015maximin}, robotics \cite{FuKSA23}, and more.}
The problem has been studied for decades, yet has recently
received renewed attention as modern applications increasingly
require reasoning about multiple competing objectives
\cite{salzman2025multi}.
Over this period, substantial progress has been made in exact and approximate multi-objective search (MOS) algorithms~\cite{SalzmanF0ZCK23}.

Despite this algorithmic progress, empirical evaluation in
MOS remains fragmented: different works typically evaluate algorithms on problems drawn from different benchmark families, including road networks, synthetic graph generators, and grid-based pathfinding environments \cite{dimacs9sp,funke2017personal_routes,harabor2024guards,weise2022manyobjective_benchmark}, often using their own benchmark instances (see, e.g.,~\citealp{maristany2023targeted}).
Unfortunately, these instances often differ in objective definitions, query-generation procedures, and reporting protocols, making comparisons across different works indirect, difficult to reproduce, and sensitive to instance-specific modeling choices \cite{SalzmanF0ZCK23}. In addition, DIMACS benchmarks, 
which are among the most
commonly used datasets in MOS studies
(e.g.,~\cite{ahmadi2024nwmoa}),
have recently been shown to exhibit highly correlated objectives
\cite{halle2025correlated}, making them a questionable go-to benchmark.

Because MOS evaluation is sensitive to instance properties (e.g., number of objectives, graph connectivity, edge-cost distributions, and inter-objective correlation, which influence Pareto-front size and structure), a standard benchmark must span the full spectrum of domain structures. 
To address this challenge, this paper takes a first step toward standardization by introducing a benchmark suite unifying diverse problem families under a consistent formulation and evaluation protocol.
The suite spans four benchmark families:

\begin{itemize}
    \item[\textbf{F1}] \textbf{Road networks}, including classical DIMACS
    datasets and {multi-objective} routing instances;
    \item[\textbf{F2}] \textbf{Structured synthetic graphs}, {enabling to control the numbers of objectives and inter-objective correlation};
    \item[\textbf{F3}] \textbf{Game-based grid environments}, derived from
    weighted 2D pathfinding maps; and
    \item[\textbf{F4}] \textbf{Robot motion-planning graphs}, constructed
    from sampling-based robot  motion-planning algorithms. 
\end{itemize}
For each family, we provide fixed graph instances, standardized start-goal query sets, reference exact Pareto-optimal solution sets, and a standardized approximate evaluation protocol (Sec.~\ref{sec:pdef}). All benchmarks use non-negative, additive edge costs, compatible with classical MOS formulations.

Rather than replacing existing datasets,
the suite builds on established benchmark families while introducing new ones.
Consequently, this suite provides the missing coherent evaluation framework required for direct, reproducible comparisons of MOS algorithms. Full implementation details, generation scripts, query sets, and reference solution sets, including a step-by-step user guide for using the benchmark instances, are available in our public repository\footnote{\url{https://github.com/CRL-Technion/Multi-Objective-Search-Benchmarks}}.


\section{{Notations} and Problem Definition}
\label{sec:pdef}

We briefly formalize the exact and approximate multi-objective shortest-path (MOS) problems.
A \emph{MOS graph} is a tuple $G = (V, E, \mathbf{c})$, where $V$ is a finite
set of vertices, $E \subseteq V \times V$ is a finite set of
edges, and $\mathbf{c} : E \to \mathbb{R}^d_{\ge 0}$ assigns a
$d$-dimensional non-negative cost vector to each edge. 

A path $\pi = v_1, \dots, v_n$ is a sequence of vertices such
that $(v_i, v_{i+1}) \in E$. The \emph{cost }of $\pi$ is
$\mathbf{c}(\pi) = \sum_{i=1}^{n-1} \mathbf{c}(v_i, v_{i+1})$.


For a minimization problem, a vector $\mathbf{p}$ \emph{dominates}
$\mathbf{q}$ (denoted $\mathbf{p} \preceq \mathbf{q}$) if
$\mathbf{p}_i \le \mathbf{q}_i$ for all $i$ and
$\mathbf{p}_j < \mathbf{q}_j$ for at least one $j$.
Similarly, given an \emph{approximation factor}
$\boldsymbol{\varepsilon} \in \mathbb{R}^d_{\ge 0}$ (a dimension-specific vector, which reduces to a single scalar applied uniformly across all objectives when $\varepsilon_1 = \dots = \varepsilon_d$), we say that
$\mathbf{p}$ $\varepsilon$-dominates $\mathbf{q}$ if
$\mathbf{p}_i \le (1 + \varepsilon_i) \cdot \mathbf{q}_i$ for all $i$ 
$\mathbf{p}_j < (1 + \varepsilon_j) \cdot  \mathbf{q}_j$ for at least one $j$.

Given start and goal vertices $v_s, v_g \in V$, a solution is
any path from $v_s$ to $v_g$. The Pareto-optimal solution set
$\Pi^*$ consists of all solutions $\pi$ such that there exists
no other solution~$\pi'$ with $\mathbf{c}(\pi') \preceq \mathbf{c}(\pi)$.
Similarly, an $\boldsymbol{\varepsilon}$-approximate Pareto set $\Pi^*_{\boldsymbol{\varepsilon}}$
is a set of solutions such that every solution in $\Pi^*$ is
$\boldsymbol{\varepsilon}$-dominated by some solution in $\Pi^*_{\boldsymbol{\varepsilon}}$.

\begin{prob}[Exact MOS.]
\label{prob:exact}
    Given a MOS graph $G = (V,E,\mathbf{c})$ and vertices $v_s, v_g \in V$,
compute the Pareto-optimal solution set $\Pi^*$.
\end{prob}

\begin{prob}[Approximate MOS.]
\label{prob:apx}
Given a MOS graph $G = (V,E,\mathbf{c})$ and vertices $v_s, v_g \in V$, and $\boldsymbol{\varepsilon} \in \mathbb{R}^d_{\ge 0}$,
compute an $\boldsymbol{\varepsilon}$-approximate Pareto set
$\Pi^*_{\boldsymbol{\varepsilon}}$. 
\end{prob}

\noindent
\textbf{Note.}
when  $\boldsymbol{\varepsilon} = \boldsymbol{0}$, Prob.~\ref{prob:apx} reduces to Prob.~\ref{prob:exact}.

{\paragraph{Objective correlation.}
To characterize the linear dependence between two objectives, we utilize the \emph{Pearson correlation coefficient}~\cite{casella2021statistical}. For any two vectors $X$ and $Y$, their Pearson correlation $\rho(X, Y) \in [-1, 1]$ quantifies the strength of their linear relationship, with $1$, $-1$, and $0$ denoting strong positive, strong negative, and no linear correlation, respectively. 

In our MOS setting, for a set of edges $E$ recall that $c_i(e)$ denotes the cost of edge $e \in E$ under objective $i$. 
Following \citet{halle2025correlated}, we evaluate the pairwise correlation between objectives $i$ and $j$ by computing the Pearson coefficient over their respective edge-cost vectors:
$\rho_{i,j} := \rho(\mathbf{c}_i, \mathbf{c}_j)$,
where $\mathbf{c}_i = \langle c_i(e) \rangle_{e \in E}$. 
Throughout this work, we rely on~$\rho_{i,j}$ (often simply denoted as $\rho$ when $i$ and $j$ are clear from the context) to report and analyze the dependencies among objective edge costs.}


\section{Related Work}
\label{sec:related_work}


\paragraph{Exact and Approximate Search Algorithms.}
The landscape of multi-objective search (MOS) has matured significantly in recent years, encompassing a broad range of exact and approximate methods (see recent surveys~\cite{SalzmanF0ZCK23,salzman2025multi}). Historically rooted in multi-labeling approaches~\cite{martins1984multiobjective}, exact MOS has evolved to include specialized bi-objective algorithms such as \boastar~\cite{HernandezYBZSKS23} alongside general frameworks such as \emoastar~\cite{RenHLFKSRC25}. As exact Pareto fronts can grow exponentially with the number of objectives, bounded-approximation methods have become increasingly important. For example, \apex~\cite{zhang2022apex} uses $\varepsilon$-dominance merging to compute compact approximate Pareto sets while preserving bounds on the quality of the Pareto set.

\paragraph{Extensions beyond the classical formulation.}
The MOS community extended the ``classical'' MOS formulation (Sec.~\ref{sec:pdef}) to more general
settings such as (i)~negative edge weights~\cite{ahmadi2024nwmoa}, (ii)~aggregation-based models in
which edge costs are combined by functions other than summation~\cite{peer2025aggregation}, and
(iii)~hierarchical formulations in which objectives are ordered by priority~\cite{slutsky2021hierarchical}.
Importantly, our benchmark suite targets the ``classical'' formulation which is a deliberate scoping
choice: each such setting calls for dedicated algorithms, and we leave the corresponding benchmark
families to future releases.

\paragraph{Benchmarking and evaluation practice.}
MOS evaluation has historically relied on heterogeneous benchmarks. Real-world road networks, particularly DIMACS~\cite{dimacs9sp}, are the primary testbed for their scale and practical relevance, while high-dimensional routing models~\cite{funke2017personal_routes} and synthetic graph generators~\cite{raith2009comparison,maristany2023targeted} support the evaluation of transportation algorithms and controlled studies of objective dimensionality and interaction, respectively.

However, other search domains lack standardized MOS datasets. Grid-based pathfinding and robotic motion planning rely on standardized single-objective benchmarks such as the Moving AI repository~\cite{sturtevant2012benchmarks}, GUARDS~\cite{harabor2024guards}, and continuous motion-planning suites~\cite{moll2015benchmarking}, but standardized discrete MOS graphs for these environments are absent. Moreover, existing many-objective pathfinding suites~\cite{weise2022manyobjective_benchmark} incorporate history-dependent, non-additive costs departing from the classical formulation. Our suite addresses this gap by combining established road networks and synthetic generators with new classical MOS instances derived from these single-objective grid and motion-planning domains.

\ignore{
\paragraph{Benchmarking and evaluation practice.}
Empirical evaluation of MOS algorithms has historically
relied on heterogeneous instance sources. Real-world road
networks, particularly the DIMACS datasets, are widely used
due to their scale and realism. Synthetic graph generators
enable controlled analysis of objective dimensionality and
interaction \cite{raith2009comparison,maristany2023targeted}.
Grid-based benchmarks are common in pathfinding research,
including weighted grid environments such as GUARDS
\cite{harabor2024guards}.

High-dimensional routing models with rich edge-cost
definitions have also been explored in transportation
contexts \cite{funke2017personal_routes}, motivating evaluation in
many-objective regimes.

Weise and Mostaghim proposed a scalable many-objective
pathfinding benchmark suite used primarily in evolutionary
optimization research \cite{weise2021manyobjective_benchmark}.
Their benchmark includes exact Pareto fronts but incorporates
a history-dependent smoothness objective in which the cost
of traversing an edge depends on the previously traversed
edge. This departs from the classical edge-local cost
formulation assumed in most multi-objective shortest-path
models. Consequently, our benchmark suite focuses
exclusively on instances with edge-local cost definitions.}


\begin{table*}[t]
\centering

\caption{
%
Summary of benchmark instances and exact Pareto-set statistics,
computed using \nwmoa. \emph{Timeout} reports the number of queries that reached the 24-hour
timeout; Pareto cardinalities and runtimes are computed over completed queries only.
Grid statistics are aggregated over multiple instances of varying sizes. Large counts,
runtimes of 1{,}000s or more, and mean/median cardinalities are rounded; exact values are
available in the repository.
}

\label{tab:benchmark-summary}
\footnotesize
\resizebox{\textwidth}{!}{%
\begin{tabular}{l l r r c r r r r r r}
\toprule
\multirow{2}{*}{Family} & \multirow{2}{*}{Benchmark} & \multirow{2}{*}{Vertices} & \multirow{2}{*}{Edges} & \multirow{2}{*}{Timeout} & \multicolumn{4}{c}{$|\Pi^*|$} & \multirow{2}{*}{Med.\ (s)} & \multirow{2}{*}{Max\ (s)} \\
\cmidrule(lr){6-9}
& & & & & Min & Max & Med. & Mean & & \\
\midrule

\multirow{6}{*}{\textbf{F1}}
& DIMACS-NY\_2 & 264{,}346 & 733{,}846 & 0/100 & 3 & 808 & 94 & 157 & 0.04 & 0.90 \\
& DIMACS-FLA\_2 & 1{,}070{,}376 & 2{,}712{,}798 & 0/100 & 2 & 3{,}856 & 241 & 511 & 0.16 & 17.52 \\
& DIMACS-Extended-NY\_3 & 264{,}346 & 733{,}846 & 0/100 & 14 & 352{,}448 & 9{,}291 & 32{,}510 & 12.33 & 3.81K \\
& DIMACS-Extended-NY\_4 & 264{,}346 & 733{,}846 & 31/100 & 24 & 1{,}226{,}887 & 47{,}458 & 167{,}685 & 75.94 & 29.2K \\
& DIMACS-Extended-NY\_5 & 264{,}346 & 733{,}846 & 40/100 & 24 & 1{,}695{,}842 & 69{,}567 & 212{,}460 & 63.89 & 17.6K \\
& GER\_10 & 10{,}000 & 19{,}164 & 3/100 & 0 & 211{,}259 & 2{,}734 & 15{,}425 & 29.98 & 31.9K \\
\midrule

\multirow{5}{*}{\textbf{F2}}
& Grid\_2$_{k \times m}$ & $km + 2$ & $\sim 2km$ & 0/49 & 329 & 779 & 563 & 559 & 3.74 & 16.72 \\
& Grid\_3$_{k \times m}$ & $km + 2$ & $\sim 2km$ & 0/49 & 42{,}087 & 313{,}847 & 139{,}474 & 153{,}655 & 1.28K & 2.79K \\
& NetMaker-10k\_3 & 10{,}000 & $\sim$60K & 0/200 & 36 & 1{,}343 & 252 & 326 & 0.03 & 0.25 \\
& NetMaker-20k\_3 & 20{,}000 & $\sim$120K & 0/200 & 67 & 1{,}276 & 368 & 424 & 0.06 & 0.57 \\
& NetMaker-30k\_3 & 30{,}000 & $\sim$180K & 0/200 & 101 & 1{,}423 & 418 & 485 & 0.09 & 0.77 \\
\midrule

\textbf{F3}
& Games-GUARDS\_2 & 163K--1.05M & 408K--5.99M & 0/1200 & 0 & 807 & 90 & 118 & 0.18 & 53.77 \\
\midrule

\multirow{2}{*}{\textbf{F4}}
& Panda-RRG\_2 & 20{,}000 & 364{,}770 & 0/100 & 2 & 258 & 50 & 65 & 0.06 & 0.30 \\
& Panda-RRG\_8 & 1{,}000 & 10{,}260 & 9/100 & 30 & 1{,}161{,}690 & 107{,}825 & 165{,}744 & 1.23K & 22.2K \\
\bottomrule
\end{tabular}%
}
\\
$k,m \in \{300,350,400,450,500,550,600\}$
\end{table*}




\section{The Benchmark Suite}
We present our benchmark suite for MOS evaluation. We first describe the design goals guiding its construction (Sec.~\ref{subsec:design-goals}), then introduce its four families: existing road-network and synthetic-graph benchmarks, and new game-based grid and robot motion-planning benchmarks. For each family, we specify the objective definitions, graph structure, and  query sets (Sec.~\ref{subsec:road}--\ref{subsec:mp}). Table~\ref{tab:benchmark-summary} summarizes the scale (vertex and edge counts), exact Pareto-front statistics, and exact-search runtime statistics of all benchmark instances. Throughout, a benchmark name's trailing number (e.g., \texttt{GER\_10}) denotes its objective count.

All benchmark graphs are directed; when the underlying domain is undirected, edges are modeled bidirectionally.

\subsection{Benchmark Design Goals}
\label{subsec:design-goals}

The benchmark suite is designed to support meaningful and reproducible empirical evaluation of MOS algorithms. Its construction is guided by the following goals.

\begin{itemize}
    \item[\textbf{G1}] 
    %

    \textbf{Standardization.} Benchmarks are defined using fixed graph instances, objective formulations, and the same publicly released start--goal query sets, enabling direct comparison across algorithms.
        
    \item[\textbf{G2}] 
    \textbf{Reproducibility.} All benchmark components, including objective definitions, query sets, and reference solution sets, are fully documented and deterministic.
    We provide all source code and graph-generation scripts to allow results to be independently verified and to enable future users to generate additional benchmark instances.

    \item[\textbf{G3}] 
    \textbf{Diversity.} The suite includes instances with varying numbers of objectives and objective interactions. These span qualitatively different graph topologies and cost-generation mechanisms, ranging from strongly correlated to nearly independent settings. {As we will see, some benchmark families are designed to contain large Pareto fronts, while others include instances with disconnected components, requiring algorithms to efficiently determine that the start and goal are disconnected.} Finally, we randomly select queries in each domain and avoid curating queries toward a target difficulty, as doing so risks selecting pathological outliers rather than reflecting the domain's natural difficulty distribution.
    
    \item[\textbf{G4}]
    \textbf{Compatibility.} When possible, our  suite builds on existing graph datasets and established construction methodologies, even when alternative approaches are possible, to ensure consistency with prior work.

    \item[\textbf{G5}] 
    \textbf{Extensibility.} The benchmark suite, alongside the provided generation scripts, is structured to facilitate the community-driven addition of new benchmark families, objectives, and query sets as MOS research evolves.
\end{itemize}

\subsection{Road Networks}
\label{subsec:road}

Road networks form a central benchmark family in MOS due to their practical relevance and widespread use in routing and transportation problems. While they are commonly used to evaluate routing metrics such as distance and travel time, these classical objective pairs often exhibit near-perfect correlation (i.e., $\rho \sim 1$), severely limiting the resulting trade-offs and Pareto-front cardinality \cite{halle2025correlated}. To address this and capture meaningful objective conflicts, the benchmark suite includes multiple datasets that extend beyond standard correlated pairs, varying in graph size, objective dimensionality, and objective definitions.

\subsubsection{\texttt{DIMACS} Road Networks}

\paragraph{Objectives.}
The \texttt{DIMACS} road-network benchmark defines two objectives: geographic distance and travel time. Both objectives are derived from the original DIMACS datasets and accumulated additively along paths, following the  formulations used in the 9th DIMACS Implementation Challenge \cite{dimacs9sp}.

\paragraph{Structure.}
We include the New York (NY) and Florida (FLA) road networks. 

\paragraph{Objective interaction.}
Distance and travel time are strongly positively correlated at the edge level, indicating substantial redundancy between the two classical objectives. Specifically, $\rho = 0.96$ in NY and $\rho = 0.97$ in FLA.

\paragraph{Queries.}
Each instance provides 100 start--goal queries, sampled uniformly at random.

\subsubsection{\texttt{DIMACS-Extended} Road Networks}

\paragraph{Objectives.}
The \texttt{DIMACS-Extended} benchmark augments the NY road network with additional objectives: { absolute elevation difference between two adjacent vertices, average vertex outdegree of adjacent vertices, and hop count (unit edge cost), }in addition to distance and travel time. Objective definitions follow the methodology introduced in the NWMOA* road-network benchmark \cite{ahmadi2024nwmoa}.

\paragraph{Structure.}
All variants are constructed on top of the NY graph, preserving its topology while extending dimensionality to three, four, or five objectives.

\paragraph{Objective interaction.}
While distance and travel time remain strongly correlated ($\rho \approx 0.96$), elevation exhibits only moderate correlation with them ($\rho = 0.38$ for both). Furthermore, average vertex outdegree and hop count exhibit negligible correlation with all other metrics (all $|\rho| < 0.08$), successfully introducing partially independent objective dimensions into the road network topology.

\paragraph{Queries.}
To enable direct performance comparisons against the base bi-objective setting, all extended variants evaluate on the same 100 fixed start--goal queries defined for the base (non-extended) \texttt{DIMACS-NY\_2} instance above.

\subsubsection{GER10 Road Network}
\paragraph{Objectives.}
The \texttt{GER\_10} benchmark defines ten edge-cost objectives following the high-dimensional routing objective definitions introduced in \cite{funke2017personal_routes}, including distance, travel time, ascent, large roads, medium roads, small roads, gas price, energy, unit, and quietness. All objectives are represented as separate edge-cost functions and accumulated additively along paths.

\paragraph{Structure.}

The benchmark uses a 10{,}000-vertex subgraph of the original German road network (approximately 4.8 million vertices), extracted by breadth-first search (BFS) from a fixed, arbitrarily chosen source vertex. This size keeps the benchmark computationally tractable, particularly given \texttt{GER\_10}'s high objective dimensionality, while remaining sufficiently large to produce challenging MOS instances. As the underlying  network contains one-way streets, the extracted subgraph is not necessarily strongly connected, and a  few sampled query pairs are disconnected in the reverse direction (see \emph{Queries} below). Finally, vertices are remapped to contiguous identifiers while preserving spatial locality.

\paragraph{Objective interaction.}
\texttt{GER\_10} has multiple highly correlated objective pairs. Routing-related objectives (e.g., distance, energy consumption, gas price, and small-road preference) exhibit strong positive correlation ($\rho > 0.95$). Other pairs, including distance--travel time and road-type interactions, exhibit moderately strong correlation ($\rho \approx 0.8$).

\paragraph{Queries.}
Consistent with the other road-network instances, evaluation is standardized over a fixed set of 100 randomly sampled start--goal pairs. A small number of these pairs (3 out of 100) are disconnected due to one-way streets in the underlying network; such cases are retained and result in an empty Pareto-optimal solution set, consistent with our handling of disconnected queries elsewhere in the suite (Sec.~\ref{subsec:game}).

\subsection{Synthetic Graphs}
\label{subsec:synthetic}

Synthetic graph benchmarks provide a controlled environment for systematic experimentation. Their structure and cost assignments are generated programmatically, enabling precise control over graph topology, number of objectives, and objective interactions. In this suite, we include both independently generated grids and structured synthetic graph models (to be explained shortly). 

\subsubsection{Random-Cost Grids}

\paragraph{Objectives.}
The \texttt{Grid} benchmark assigns either two or three objective costs to grid edges.
Each objective is generated independently by sampling integer costs uniformly at random
from $[1,10]$, following the methodology of \citet{maristany2023targeted} and~\citet{de2021improved}.

\paragraph{Structure.}
The benchmark contains two-dimensional four-connected grid graphs constructed by independently selecting  the width and height from
$\{300,350,400,450,500,550,600\}$, yielding a total of $7 \times 7 = 49$ different graphs.
These graphs are shared for each number of objectives ($2$ and $3$).

Each instance includes two auxiliary vertices defining a
source--target query: source and target vertices connected to all vertices in the leftmost and rightmost columns, respectively. The auxiliary edges draw costs from the same random distribution as all other edges.

\paragraph{Objective interaction.}
Because costs are sampled uniformly and independently, objectives are nearly independent by construction. Across all variants, $|\rho| < 0.01$.

\paragraph{Queries.}
Each instance defines a single source--target query between the
auxiliary source and target vertices.

\subsubsection{NetMaker}

\begin{figure}[t]
\centering
\includegraphics[width=0.48\linewidth]{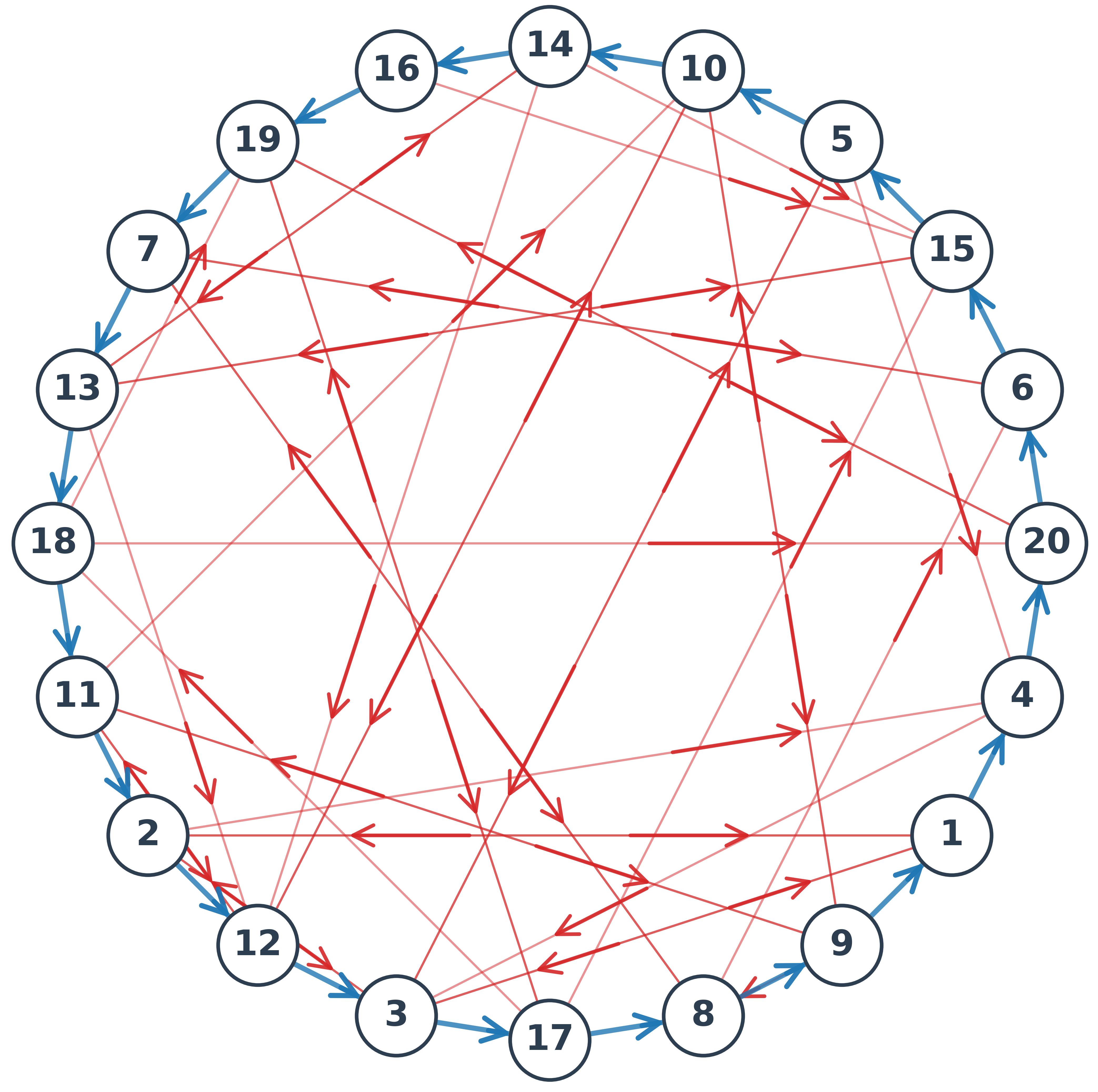}
\caption{
Illustration of the \texttt{NetMaker} graph construction 
for  \(\vert V \vert=20\), \(a_{\min}=1\), \(a_{\max}=4\), and~\(I_{\text{vertex}}=4\). 
Blue and red edges correspond to 
$E_{\mathrm{cyc}}$ and~$E_{\mathrm{loc}}$, respectively.
Vertices are arranged according to the Hamiltonian cycle
for visualization; since locality is defined in vertex-ID space rather
than cycle order, some locality edges appear to connect distant vertices
in the drawing.}
\label{fig:netmaker-structure}
\end{figure}

\paragraph{Objectives.}
\texttt{NetMaker} defines three non-negative, integer-valued
objectives accumulated additively along paths, following
synthetic constructions inspired by prior MOS
benchmarks \cite{raith2009comparison,maristany2023targeted}.
Each edge $e$ has a cost vector \(\mathbf{c}(e)\in\mathbb{Z}_{\ge 0}^3\).

\paragraph{Structure.}
Each instance is a directed graph \(G=(V,E)\) with
\(V=\{1,\dots,N\}\) and
$E = E_{\mathrm{cyc}} \cup E_{\mathrm{loc}}$.
To define \(E_{\mathrm{cyc}}\), a uniformly  random permutation
\(\sigma=(\sigma_1,\dots,\sigma_N)\) of \(V\) is sampled, and the directed
Hamiltonian cycle
\[
E_{\mathrm{cyc}}
=
\{(\sigma_i,\sigma_{i+1}) \mid 1 \le i < N\}
\cup
\{(\sigma_N,\sigma_1)\}
\]
is added. This guarantees that every vertex lies on a directed cycle
connecting all vertices, ensuring strong connectivity.

The additional edge set \(E_{\mathrm{loc}}\) controls graph density while introducing local routing alternatives, whereas the Hamiltonian cycle preserves global connectivity. Let \(a_{\min}\) and \(a_{\max}\) denote the minimum and maximum out-degree parameters. For each vertex \(u \in V\), a target out-degree is sampled uniformly from \(\{a_{\min}+1,\dots,a_{\max}\}\) (the extra \(1\) accounts for the cycle edge already present), and edges are added within the locality window \(W(u)\), where \(I_{\text{vertex}}\in\mathbb{Z}^+\) defines the locality window size:
\[
W(u)=\Big[\max\big(1,u-\lfloor\tfrac{I_{\text{vertex}}}{2}\rfloor\big),
\min\big(N,u+\lfloor\tfrac{I_{\text{vertex}}}{2}\rfloor\big)\Big],
\]
excluding self-loops, duplicate edges, and edges already in~\(E_{\mathrm{cyc}}\), until the sampled out-degree is reached or no valid local edge remains. Locality is defined in vertex-ID space rather than along the Hamiltonian cycle (Fig.~\ref{fig:netmaker-structure}); increasing~\(I_{\text{vertex}}\)  produces instances with different routing difficulty and Pareto-front structures.

Costs depend on edge type. For each cycle edge
\(e\in E_{\mathrm{cyc}}\), exactly one objective value is sampled from
each of the intervals \([1,333]\), \([334,666]\), and \([667,1000]\),
with the assignment of intervals to objectives randomly
permuted independently per edge. For each additional edge
\(e\in E_{\mathrm{loc}}\), all three objective values are sampled
independently and uniformly from \([1,99]\). This makes shortcut edges cheaper than
cycle edges in all objectives, creating alternative routes while
the Hamiltonian backbone sets the overall path length scale.

The current release contains 12 graphs: three size groups
(\(N\in\{10{,}000,20{,}000,30{,}000\}\)), two locality settings
(\(I_{\text{vertex}}\in\{20,50\}\)), and two graphs per configuration.
All instances use \(a_{\min}=1\) and \(a_{\max}=10\), yielding
average out-degree close to \(6\). Table~\ref{tab:benchmark-summary} reports the average
edge counts for the three size groups. Fig.~\ref{fig:netmaker-structure} illustrates the construction.

\paragraph{Objective interaction.}
Objective interaction differs by edge type. On
Hamiltonian-cycle edges ($E_{\mathrm{cyc}}$), each edge receives one low,
medium, and high magnitude cost component, randomly
assigned to the three objectives. This induces moderate
negative correlation between objective pairs
(\(\rho \approx -0.44\) on average). In contrast, additional
locality edges ($E_{\mathrm{loc}}$) are assigned independent uniform costs in
\([1,99]\) and therefore exhibit negligible correlation
(\(\rho \approx 0\)).

\paragraph{Queries.}
Each graph is paired with 50 randomly sampled start–goal
queries. Source and target vertices are sampled from the first and last~\(10\%\)
of vertex identifiers, respectively.
As additional edges are restricted by the locality
parameter~\(I_{\text{vertex}}\), most paths must traverse
multiple graph regions rather than
using direct  shortcuts. This encourages
alternative paths,
resulting in large Pareto fronts while reducing trivial or very small Pareto fronts. Other families in the suite capture a broader range of Pareto-front sizes.

\subsection{Game-Based Grid Environments}
\label{subsec:game}
Game-based grid benchmarks are derived from weighted pathfinding maps designed for video-game
environments, which exhibit diverse geometric layouts including open regions, mazes, and
room-based layouts. This family complements the road-network and synthetic families with
realistic layouts and structurally independent objectives.

\ignore{
Game-based grid benchmarks are derived from weighted pathfinding maps originally designed for single-objective evaluation in video-game environments. These maps exhibit diverse geometric layouts, including open regions, mazes, and room-based layouts.}

\ignore{
Each map is converted into a bi-objective shortest-path instance with objectives of path length and guard exposure (guard count with line-of-sight to a cell). The resulting large, 8-connected graphs exhibit weak objective correlation, complementing road-network and synthetic benchmarks with realistic layouts and structurally independent objectives.}

\begin{figure}[t]
    \centering
    \begin{subfigure}[t]{0.40\linewidth}
        \centering
        \includegraphics[width=0.85\linewidth]{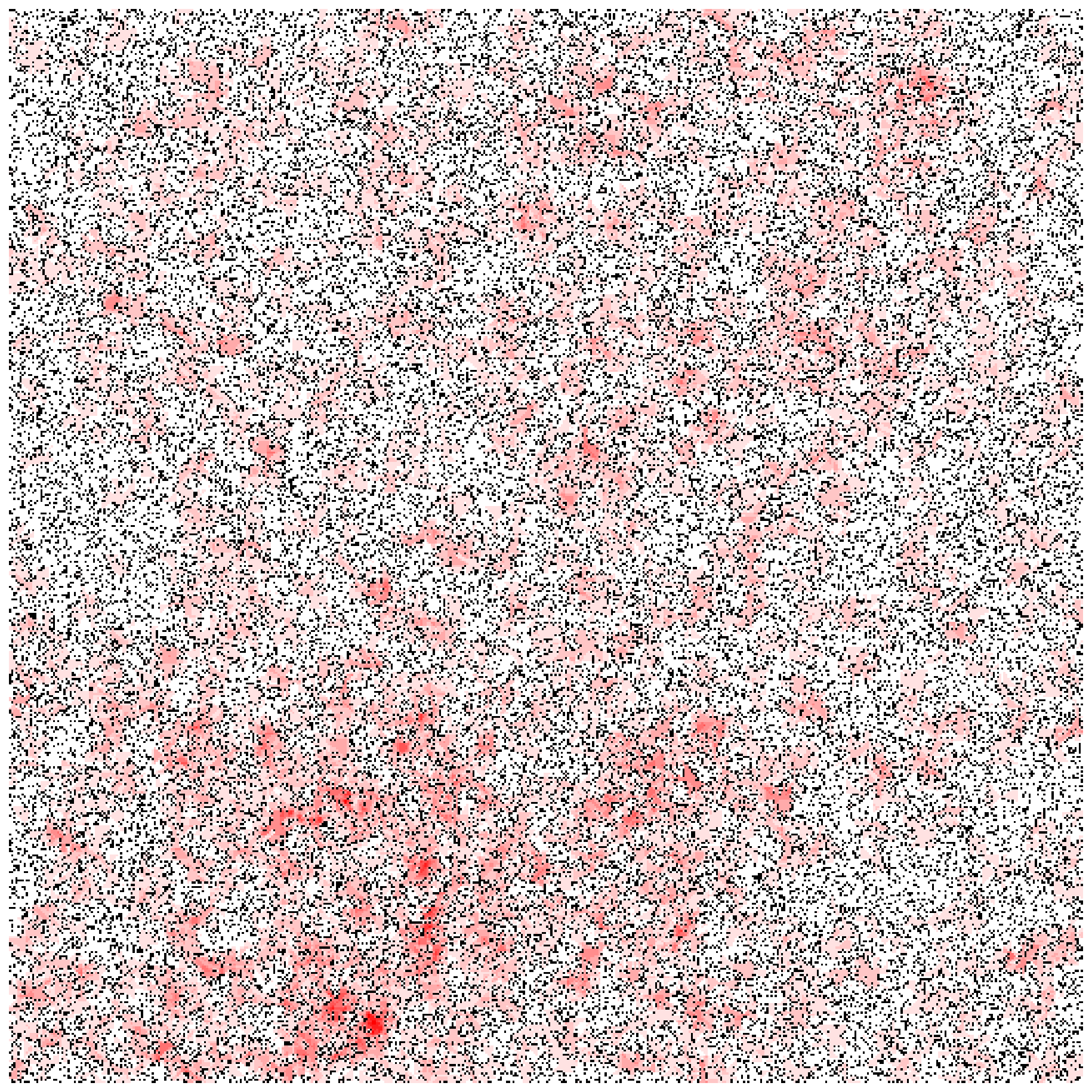}
        \caption{random512-20-1.}
        \label{fig:random-map}
    \end{subfigure}
    \hfill
    \begin{subfigure}[t]{0.40\linewidth}
        \centering
        \includegraphics[width=0.85\linewidth]{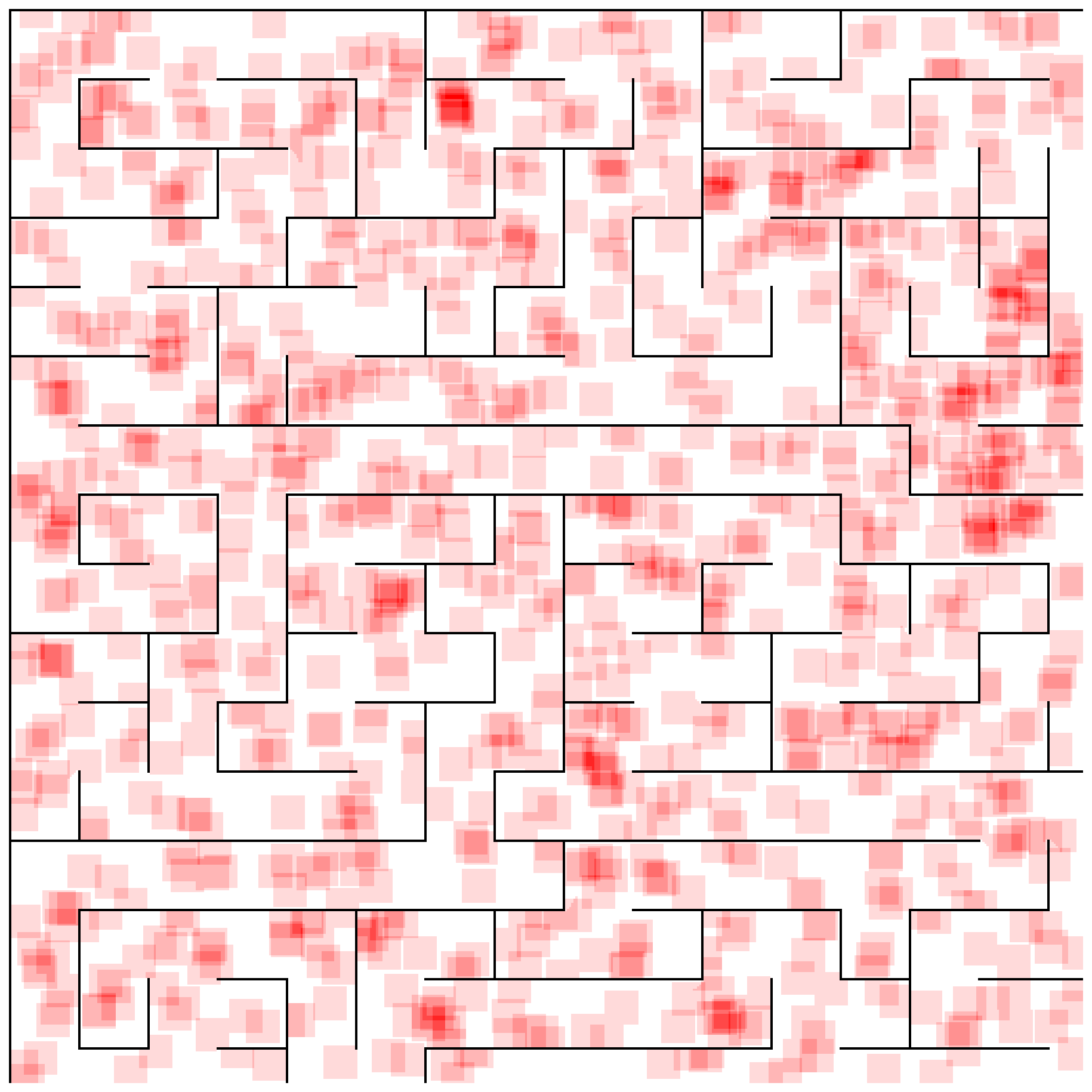}
        \caption{Maze512.}
        \label{fig:maze-map}
    \end{subfigure}

    \caption{
    Two representative \texttt{Games} benchmark environments (of 12 total).
    Black regions denote obstacles. Red intensity reflects cost magnitude with darker red indicating regions observed by multiple guards.
    }
    \label{fig:grid-benchmarks}
\end{figure}

\subsubsection{GUARDS-Based Grid Benchmarks}




\paragraph{Objectives.}
The \texttt{Games} benchmark converts each map of the single-objective GUARDS
 pathfinding benchmark~\cite{harabor2024guards} into bi-objective instances whose 
objectives are path length and guard exposure. Path length assigns costs of 10 and 14 to orthogonal
and diagonal moves, respectively, approximating Euclidean distance.

\ignore{The \texttt{Games} benchmark defines two objectives on grid maps derived from the GUARDS weighted pathfinding benchmark~\cite{harabor2024guards}. 
While the original GUARDS benchmark was designed for single-objective weighted pathfinding, we convert each map into a bi-objective shortest-path formulation with objectives of path length and guard exposure. Path length assigns costs of~10 and~14 to orthogonal and diagonal moves, respectively, approximating Euclidean distance.}

Guard exposure captures spatially structured risk. Each cell is assigned the maximum number of guards with line-of-sight to it. The exposure cost of an orthogonal move equals the destination cell's value, while for a diagonal move it equals the maximum value among the destination cell and the two intermediate side cells (diagonal moves are allowed only when both side cells are passable). Both objectives are non-negative and additive, with path exposure defined as the sum of per-move exposure costs, not a per-cell maximum.

\paragraph{Structure.}
Each map is converted into a directed eight-connected graph, where each valid move becomes a directed edge.
The current release includes 12 large-scale GUARDS maps; two representative examples appear in Fig.~\ref{fig:grid-benchmarks}.

\paragraph{Objective interaction.}
Across the 12 GUARDS maps, the Pearson correlation between objectives remains small, with mean $\rho = 0.062$ and maximum $\rho = 0.123$ (observed for \texttt{random512-20-1}), indicating near independence.

\paragraph{Queries.}
For each map, 100  start–goal queries are defined. 
Queries are sampled uniformly from passable cells, with some corresponding to disconnected source–target pairs; such cases are retained and result in empty solution sets.

\subsection{Robot Motion-Planning Roadmaps}
\label{subsec:mp}

\begin{figure}[t]
    \centering

    \begin{subfigure}[t]{0.40\linewidth}
        \centering
        \includegraphics[width=0.97\linewidth]{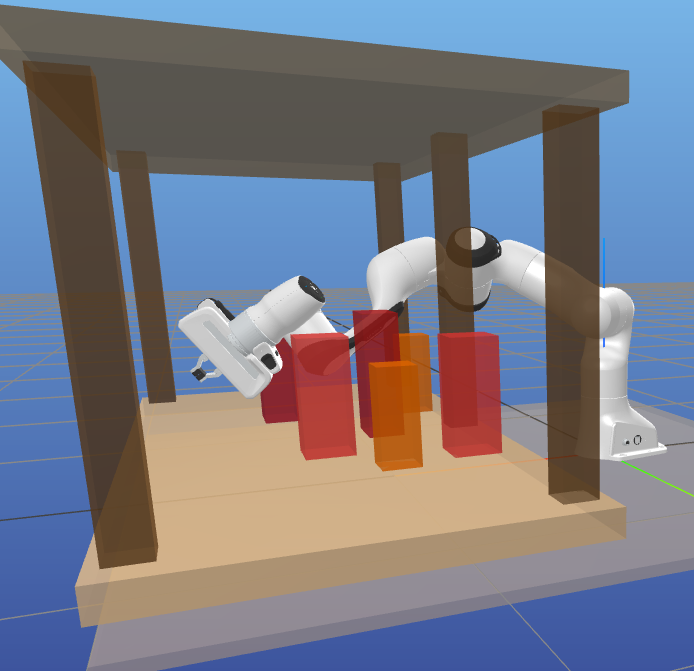}
        \caption{}
        \label{fig:panda-prm-environment}
    \end{subfigure}
    \hfill
    \begin{subfigure}[t]{0.40\linewidth}
        \centering
        \includegraphics[width=0.97\linewidth]{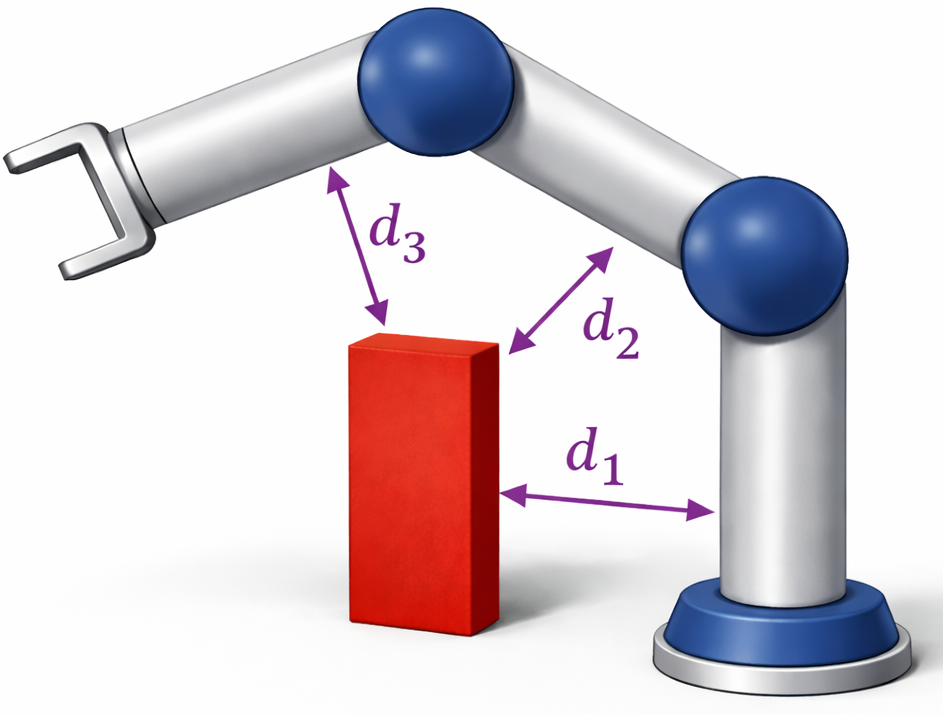}
        \caption{}
        \label{fig:per_link_clearance}
    \end{subfigure}

    \caption{
     (a) RRG benchmark environment of a Franka Emika Panda manipulator in a tabletop scene featuring vertical support pillars and box-shaped obstacles. (b) Illustration of link-specific clearance distances ($d_1$, $d_2$, $d_3$), 2D setting. Unlike an aggregated clearance model collapsing safety into a minimum-distance penalty (here, $d_2$), a per-link formulation preserves the geometric risk profile across the manipulator.
    }
    \label{fig:panda_benchmark}
\end{figure}

Robot motion-graph benchmarks are derived from sampling-based manipulation planning problems.
Roadmap graphs are constructed for the Franka Emika Panda 7-DoF arm in a fixed tabletop
manipulation scene (Fig.~\ref{fig:panda-prm-environment}), complementing the suite with realistic
high-dimensional graphs whose objectives trade off motion efficiency and obstacle clearance.

\subsubsection{Panda Motion-Planning Roadmaps}

\paragraph{Objectives.} Each edge $(u,v)$ of a \texttt{Panda} roadmap corresponds to a collision-free linear
interpolation between joint configurations $q_u$ and $q_v$. The first objective is joint-space
path length, $c_1(u,v) = \|q_u - q_v\|_2$,
and the remaining objectives capture obstacle clearance, measured as the geometric minimum
distance between robot links and obstacles along the edge. Following trajectory-optimization
approaches~\cite{ratliff2009chomp,kalakrishnan2011stomp}, clearance is converted into an additive
cost by a penalty function that grows smoothly as the robot approaches an obstacle and vanishes
beyond a safety band $\delta$:
\begin{equation}
  U(d) = \begin{cases} 0 & \text{if } d \ge \delta,\\
  \frac{1}{2\delta}(d-\delta)^2 & \text{if } 0 < d < \delta, \end{cases}
\end{equation}
where $d$ denotes the distance to the nearest obstacle.

\emph{Bi-objective formulation.} Clearance is aggregated over all seven arm links,
$c_2(u,v) = U(d_{\text{min}}(u,v))$, where $d_{\text{min}}(u,v)$ is the minimum clearance over all
links along the edge, yielding \texttt{Panda-RRG\_2}.

\emph{Many-objective formulation.} Clearance is instead defined separately per link. Letting $d_i(u,v)$ denote the minimum clearance of link $i$ along the
edge, the components $c_i(u,v) = U(d_{i-1}(u,v))$ for $i = 2,\ldots,8$ yield \texttt{Panda-RRG\_8}.
Importantly, the two links closest to the base remain consistently far from obstacles, resulting
in zero clearance penalties across all edges; we retain these objectives to preserve structural
consistency across instances and to reflect the true geometric structure of the manipulator.

\paragraph{Structure.}
Roadmaps are generated using a Rapidly-exploring Random Graph (RRG) \cite{karaman2011sampling}, with a fixed connection radius of $1.5$ radians in joint space.
The clearance safety band parameter is set to $\delta = 0.1\,\text{m}$. 
The current release includes two roadmap instances: \texttt{Panda-RRG\_2} and
\texttt{Panda-RRG\_8}.

\ignore{
\emph{Bi-objective formulation.}
Clearance is computed across all seven arm links:
\[
c_2(u,v) = U\!\big(d_{\min}(u,v)\big),
\]
where $d_{\min}(u,v)$ is the minimum clearance over all links along the edge. 
This yields a two-objective problem (path length and aggregated clearance), as in instance \texttt{Panda-RRG\_2}. 

\emph{Many-objective formulation.}
Clearance is defined separately for each of the seven arm links. 
Let $d_i(u,v)$ denote the minimum clearance of link $i$ along the edge. 
For the many-objective instance, components $c_2, \dots, c_8$ correspond to per-link clearance penalties:
\[
c_i(u,v) = U\!\big(d_{i-1}(u,v)\big), 
\quad i = 2, \dots, 8.
\]
This yields eight objectives in total: joint-space path length and seven per-link clearance objectives, as in instance \texttt{Panda-RRG\_8}. In the current scene, the two links closest to the base
remain consistently far from obstacles, resulting in
zero clearance penalties across all edges. We retain these
objectives to preserve structural consistency across
instances and to reflect the true geometric structure of
the manipulator, even when certain links are weakly
constrained in a particular scene. 
}

\paragraph{Objective interaction.}
Pearson correlations are computed over all roadmap edges.
For the bi-objective instance, the correlation between path length and aggregated clearance is $\rho = 0.38$, indicating moderate positive correlation. 
For the many-objective instance, most of the 28 objective pairs exhibit weak correlation ($|\rho| < 0.3$); 
the largest are $0.48$ (between $c_4$ and $c_5$) and $0.46$ (between $c_5$ and $c_6$).

\paragraph{Queries.}
For each instance, 100 randomly sampled and fixed start--goal queries are defined.

\begin{figure}[t]
\centering

\begin{subfigure}[t]{0.48\linewidth}
\centering
\includegraphics[width=0.92\linewidth]{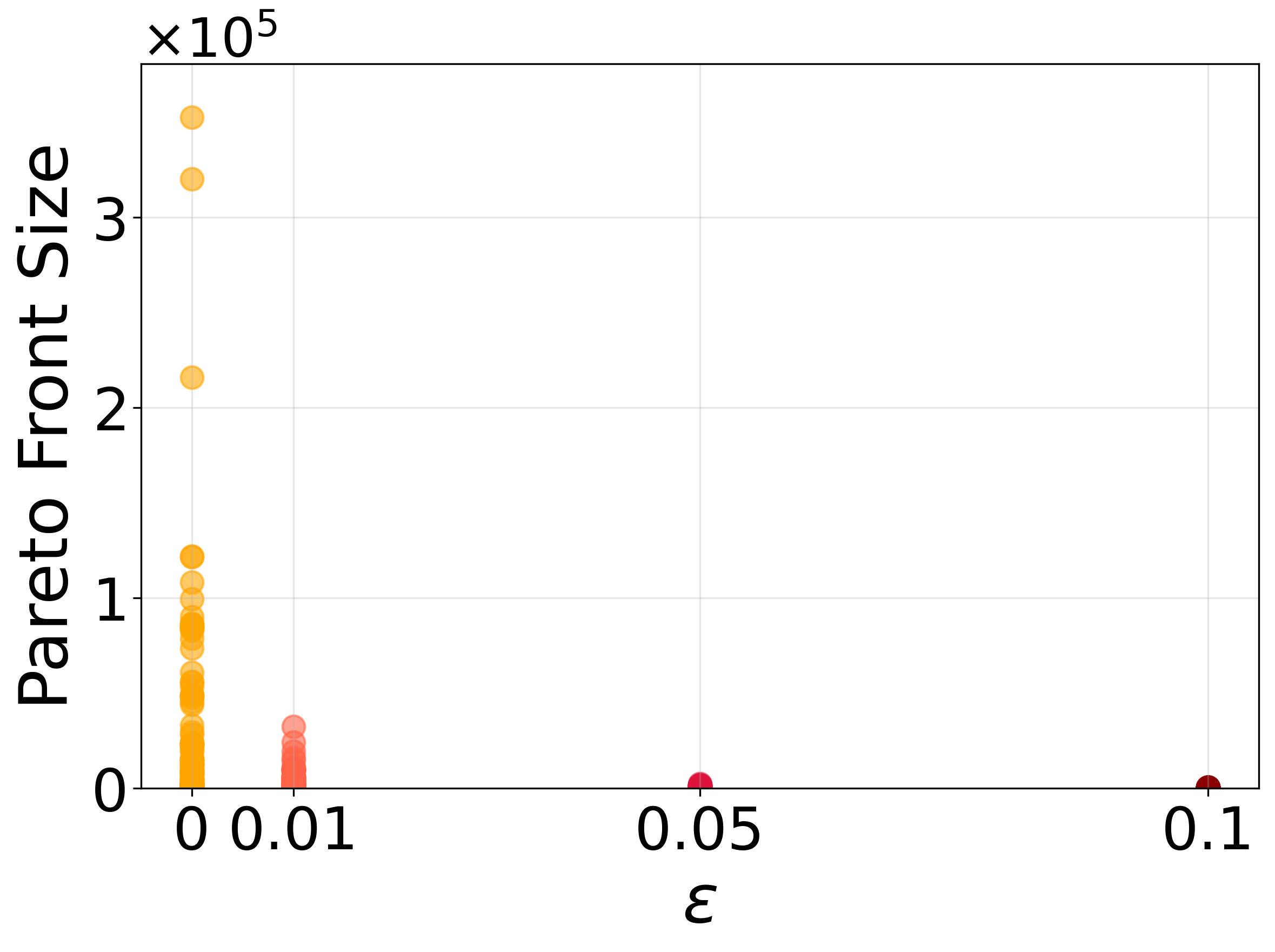}
\caption{DIMACS-Extended-NY\_3}
\end{subfigure}
\hfill
\begin{subfigure}[t]{0.48\linewidth}
\centering
\includegraphics[width=0.92\linewidth]{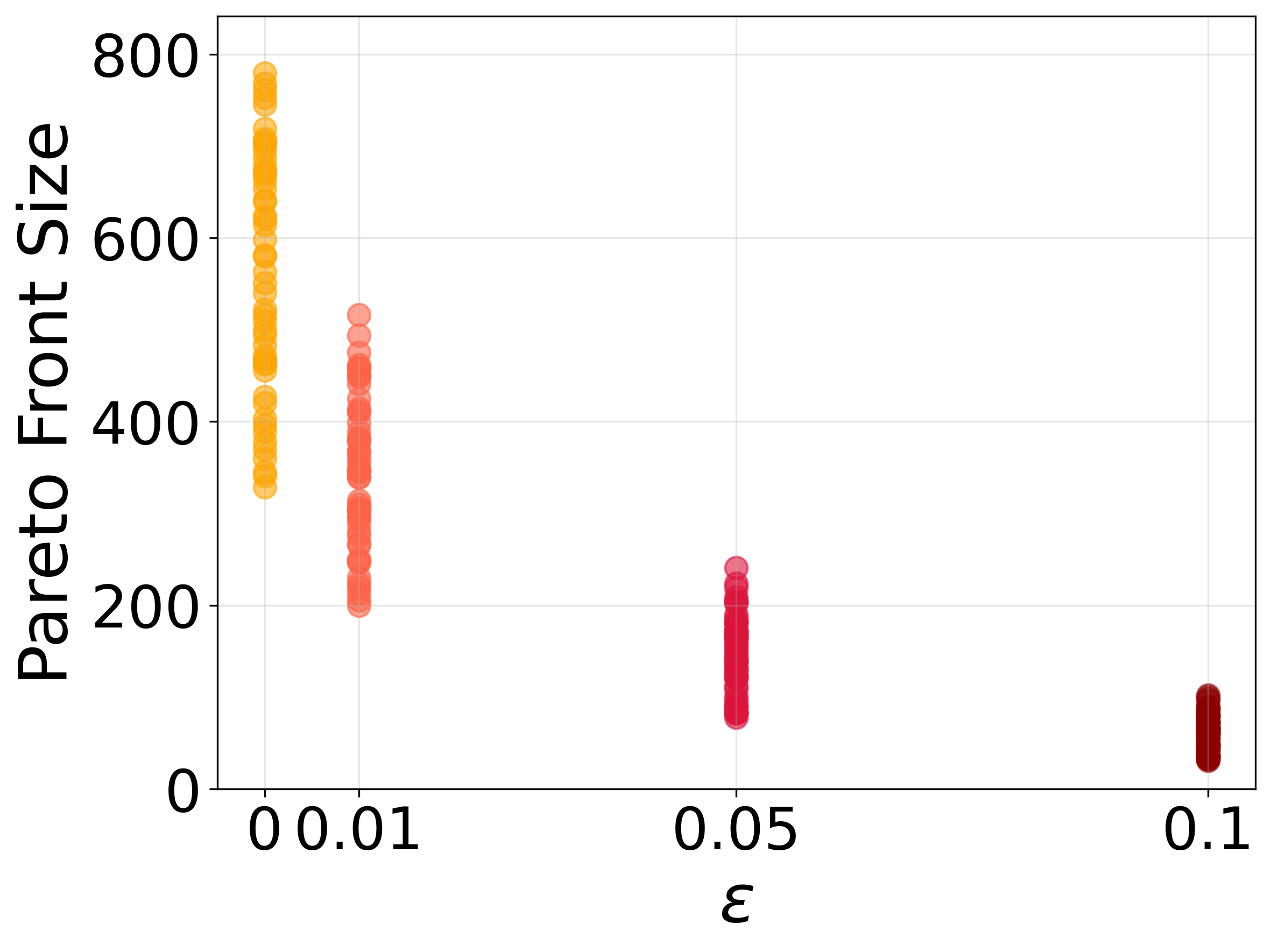}
\caption{Grid\_2$_{k \times m}$}
\end{subfigure}

\vspace{2.0mm}

\begin{subfigure}[t]{0.48\linewidth}
\centering
\includegraphics[width=0.92\linewidth]{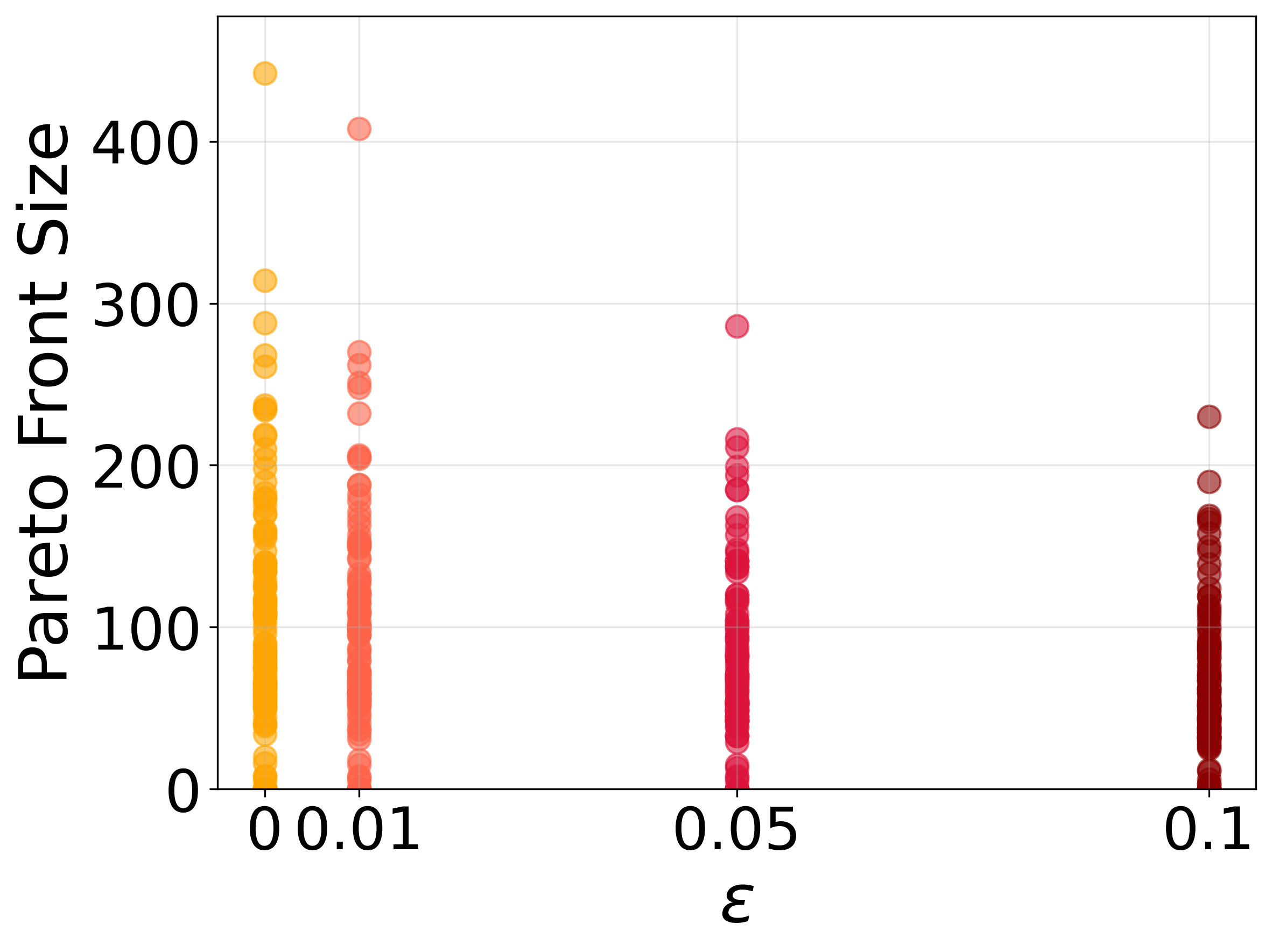}
\caption{GUARDS-DarkContinent\_2}
\end{subfigure}
\hfill
\begin{subfigure}[t]{0.48\linewidth}
\centering
\includegraphics[width=0.92\linewidth]{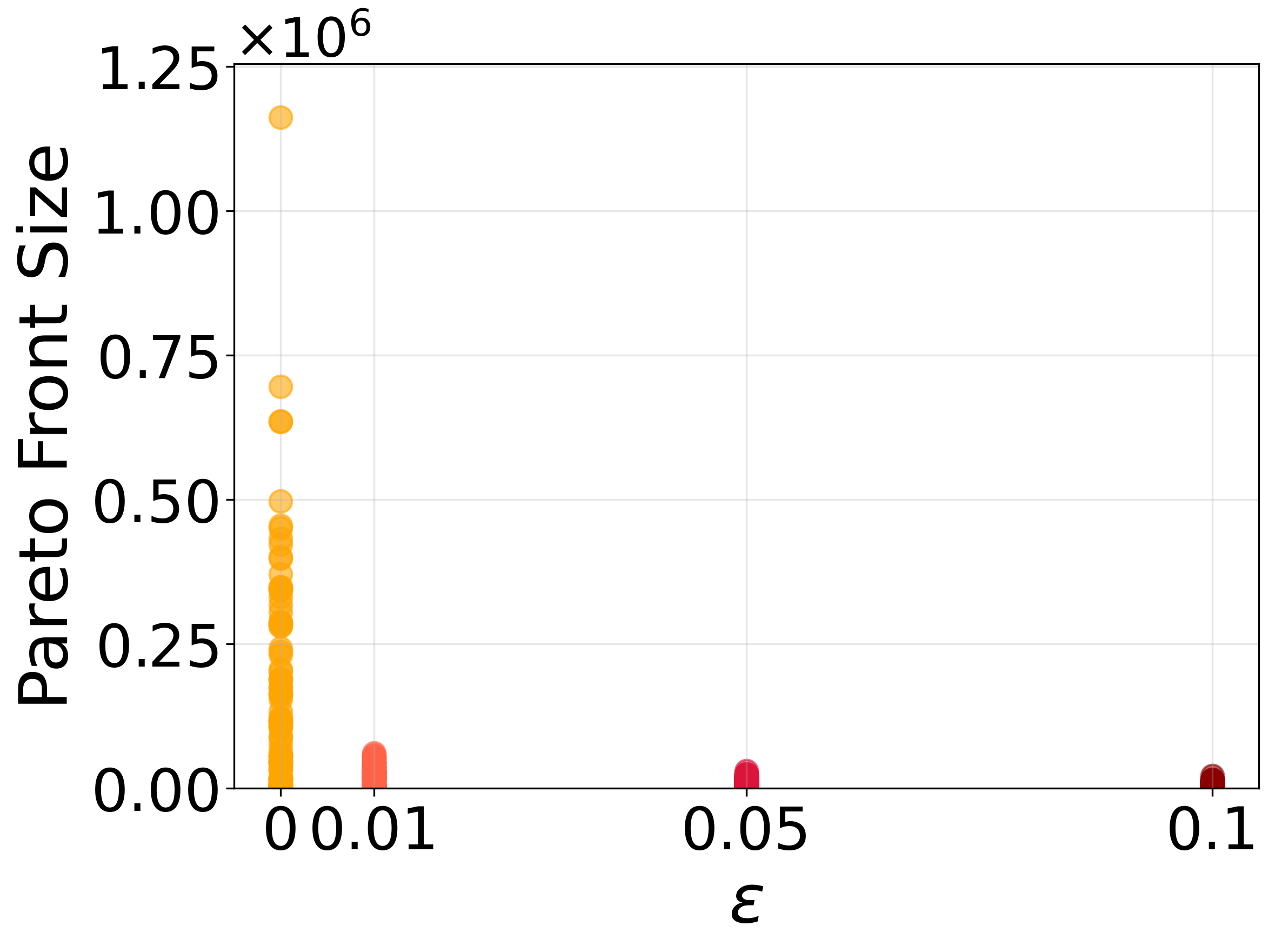}
\caption{Panda-RRG\_8}
\end{subfigure}

\caption{
Pareto-front cardinality across queries as a function of the approximation parameter $\varepsilon$ for representative benchmarks from all benchmark families.
Each point corresponds to a start–goal query.
}
\label{fig:pareto_vs_epsilon}
\end{figure}

\begin{figure}[t]
    \centering

    \begin{subfigure}[t]{0.48\linewidth}
        \centering
        \includegraphics[width=0.92\linewidth]{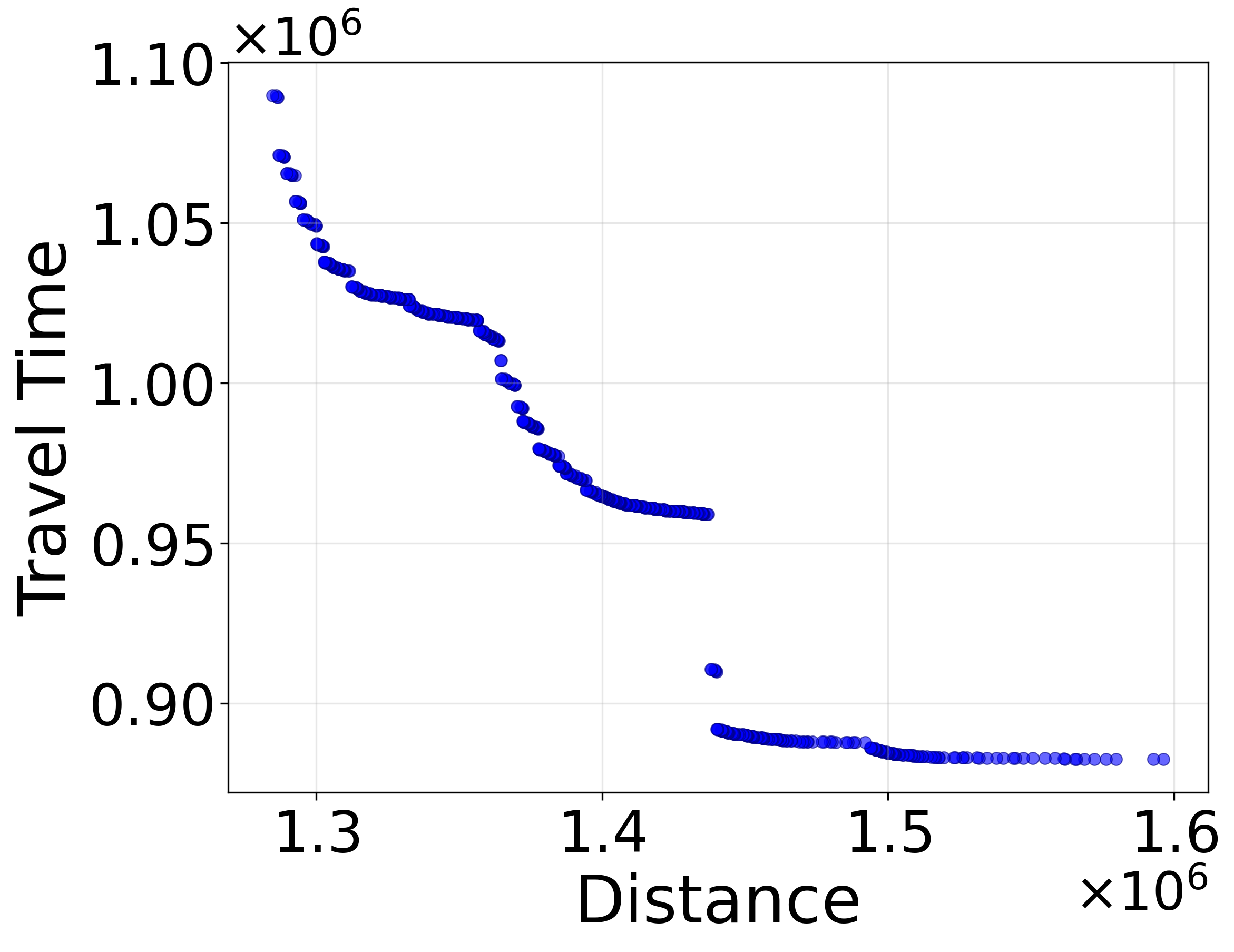}
        \caption{DIMACS-NY\_2}
    \end{subfigure}
    \hfill
    \begin{subfigure}[t]{0.48\linewidth}
        \centering
        \includegraphics[width=0.92\linewidth]{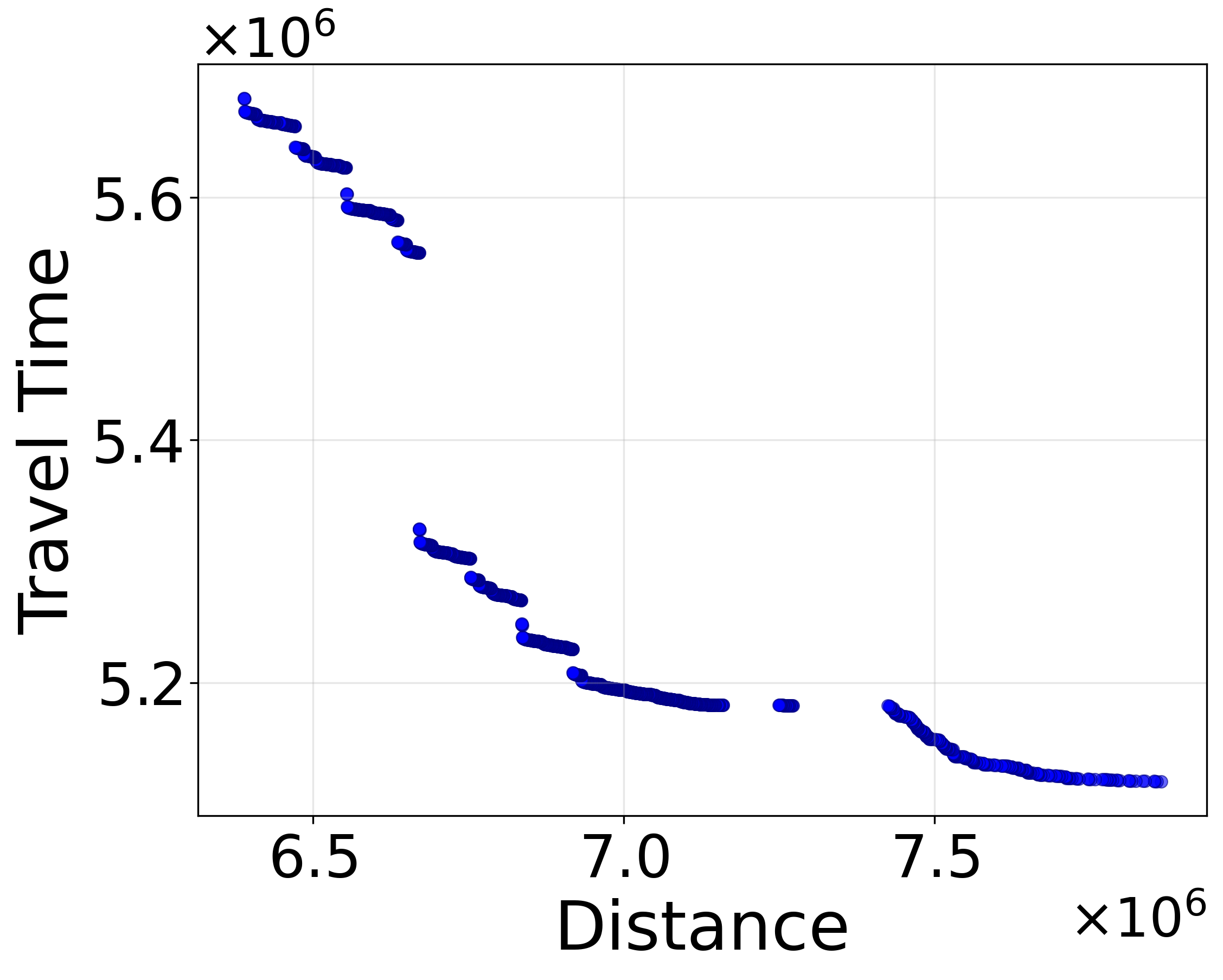}
        \caption{DIMACS-FLA\_2}
    \end{subfigure}

   \vspace{2.0mm}
   
    \begin{subfigure}[t]{0.48\linewidth}
        \centering
        \includegraphics[width=0.92\linewidth]{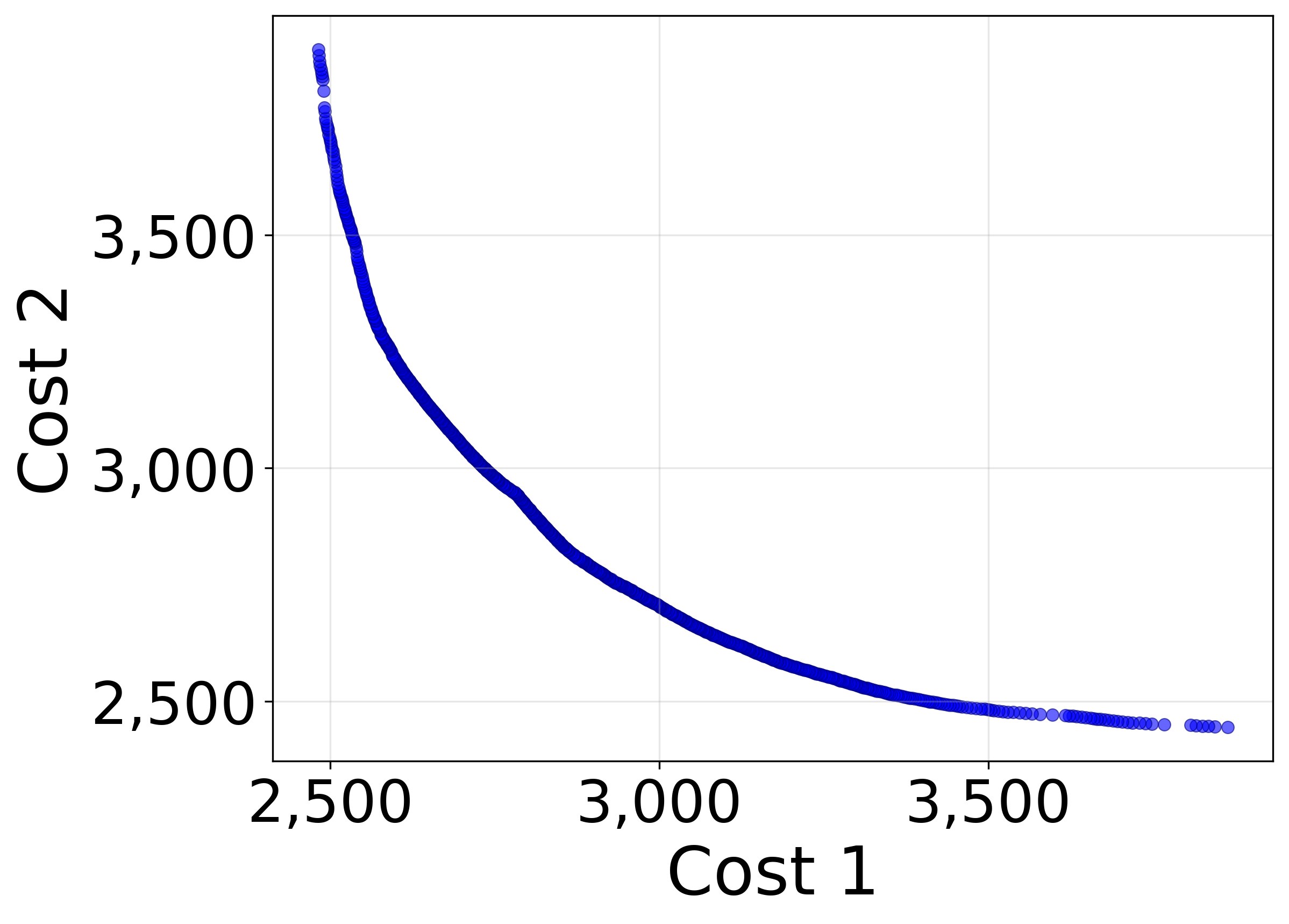}
        \caption{Grid\_2$_{k \times m}$}
    \end{subfigure}
    \hfill
    \begin{subfigure}[t]{0.48\linewidth}
        \centering
        \includegraphics[width=0.92\linewidth]{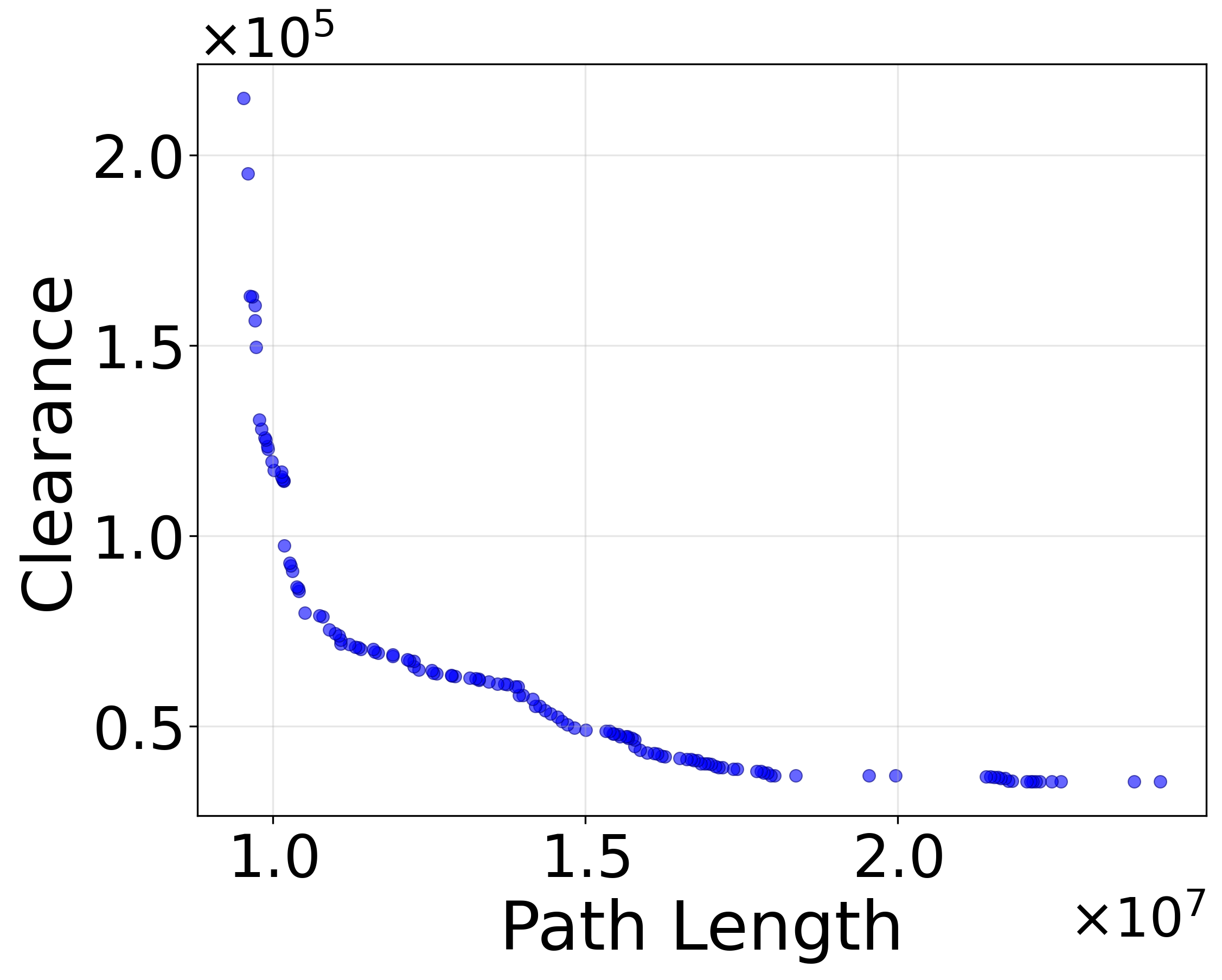}
        \caption{Panda-RRG\_2}
    \end{subfigure}

    \caption{Representative exact two-objective Pareto fronts for single queries from different benchmark families. The examples highlight substantial variation in objective-space geometry across the benchmark suite.}
    \label{fig:pareto-front-examples}
\end{figure}


\section{Standardized Evaluation Protocol}

Evaluation is performed over the fixed start--goal query sets provided for each benchmark. Exact evaluation computes the Pareto-optimal solution set for each query. Approximate evaluation uses $\varepsilon$-dominance with $\varepsilon \in \{ 0.01, 0.05, 0.1\}$, evaluating each query independently for every $\varepsilon$.


For $\varepsilon = 0$, exact reference Pareto sets are computed independently using \nwmoa~\cite{ahmadi2024nwmoa} and \ltmoa~\cite{hernandez2023lazy}, with the two implementations cross-validating one another. The repository and Tbl.~\ref{tab:benchmark-summary} report the \nwmoa solution sets for $\varepsilon = 0$. For $\varepsilon > 0$, approximate reference sets are computed using \apex~\cite{zhang2022apex} with its RANDOM merge strategy.


Runs were distributed across a shared SLURM cluster with heterogeneous compute nodes (two-socket Intel Xeon E5-2640 v3/E5-2650 v4 or AMD EPYC 7713/9654/9754 processors, providing 32--512 logical CPUs and approximately 0.5--1~TB RAM per node), running Ubuntu Linux (kernel 5.15, x86\_64). Reference-solution construction uses a uniform 24-hour per-query timeout across all benchmarks. \texttt{DIMACS-Extended} and \texttt{Grid\_3} additionally use memory caps of 275~GB and 375~GB, respectively, due to their larger working sets; no other benchmark required an explicit cap, and \nwmoa never exhausted these limits. Peak memory usage on completed queries varies by two orders of magnitude across families, from a median of 24~MB for \texttt{NetMaker} to approximately 2~GB for \texttt{Grid\_3}; even the largest single-query peaks, 6~GB on \texttt{Grid\_3} and 7~GB on \texttt{DIMACS-Extended-NY\_5}, remain well below their respective 375~GB and 275~GB caps. Tbl.~\ref{tab:benchmark-summary} reports the number of timed-out queries and \nwmoa's median and maximum runtime over completed queries. Although \ltmoa cross-validates every \nwmoa front, it is consistently slower: 2$\times$ on \texttt{DIMACS-NY\_2} (0.04s vs.\ 0.08s), 9$\times$ on \texttt{Grid\_3} ($\approx$1.2ks vs.\ $\approx$11.5s), and nearly 19$\times$ on \texttt{Grid\_2} ($\approx$3.7s vs.\ $\approx$70s). It also fails more often: on \texttt{DIMACS-Extended-NY\_5}, it times out on 49 of 100 queries versus \nwmoa's 40, and on \texttt{Grid\_3}, it exhausts memory on 28 of 49 queries while \nwmoa solves all of them. These results show that reference-solution hardness may arise from either runtime or memory demands.



Exact ($\varepsilon = 0$) reference solution sets are provided in the repository. Approximate ($\varepsilon > 0$) solution sets are used for the analyses reported here but are not distributed separately, since they are not unique benchmark artifacts: they depend on the approximation algorithm and its design choices (e.g., merge strategies in \apex), in addition to the chosen $\varepsilon$ value. Alongside each reference solution set, the repository publishes per-query \nwmoa solver statistics (runtime, expanded and generated node counts, and peak memory) for every completed query; timed-out queries are instead marked explicitly, without these counters, since they reflect an interrupted rather than a verified search.


\section{Reference Solution Properties}
Uniform random query sampling yields a broad spectrum of query hardness. Within a benchmark,
Pareto-front cardinality spans orders of magnitude (Tbl.~\ref{tab:benchmark-summary}), ranging from
2 to over 3.5K solutions for \texttt{DIMACS-FLA\_2} and from 30 to over 1.16M for
\texttt{Panda-RRG\_8}. Across benchmarks, median solver runtime similarly ranges from sub-second
(\texttt{NetMaker}, \texttt{Games-GUARDS\_2}), through tens of minutes
(\texttt{Grid\_3}, $\approx$1.3Ks), to benchmarks where many queries exhaust the 24-hour timeout
(\texttt{DIMACS-Extended-NY\_5}, \texttt{Panda-RRG\_8}). Median runtimes can mask substantial
within-benchmark variability, revealed by the gap to the maximum runtime.

Across representative benchmarks from all families, Pareto-set cardinality decreases substantially as
$\varepsilon$ increases (Fig.~\ref{fig:pareto_vs_epsilon}). Averaged over benchmark families, the
median Pareto set size at $\varepsilon = 0.1$ is reduced by roughly 90\% relative to the exact case, with per-benchmark reductions ranging from 44.6\% (\texttt{Games-GUARDS\_2}) to
99.7\% (\texttt{DIMACS-Extended-NY\_5}).

The representative Pareto fronts in Fig.~\ref{fig:pareto-front-examples} illustrate the diversity of
objective-space structures across benchmark families. For visualization, we focus on two-objective
instances, which can be directly plotted in two-dimensional objective space. Fronts differ not only in
size but also in geometry, from smooth and continuous to clustered and step-like, and in how far each
objective varies along them: measuring spread as the max/min ratio along an axis and averaging over
queries with a non-degenerate spread, both road networks stay within $1.40\times$ on either axis,
whereas on the \texttt{Games-GUARDS\_2} map \texttt{random512-20-1}, guard exposure spans
$\approx20\times$ against only $\approx2\times$ for path length. A single homogeneous benchmark could not
surface this range of structures.

\section{Conclusion and Future Work}




We presented a standardized benchmark suite for exact and approximate multi-objective search, with diverse benchmark families, standardized queries, and exact reference Pareto sets for reproducible evaluation. The suite spans road networks, synthetic graphs, game-based environments, and robotic motion-planning roadmaps, exposing objective interactions, Pareto-front structures, and computational difficulty.

This initial release is restricted to non-negative additive edge-local objectives; future work will extend the suite to aggregation-based models, hierarchical formulations, and history-dependent objectives. We anticipate that this benchmark will accelerate algorithmic development by providing a common, reproducible foundation for empirical evaluation in multi-objective search.



\bibliography{references}

\end{document}